\documentclass{article}
\usepackage{spconf,amsmath,graphicx}
\usepackage{amsmath} 
\usepackage{graphicx}
\usepackage{caption}
\usepackage{subfig, adjustbox}
\usepackage{lipsum}
\usepackage{multirow,booktabs}
\usepackage{multicol}
\usepackage{tabularx}
\usepackage{color}
\usepackage{enumitem}
\usepackage[colorlinks=True]{hyperref}
\usepackage{comment}

\title{Semantically interpretable and controllable Filter Sets}
\name{Mohit Prabhushankar*\thanks{*Equal contribution}, Gukyeong Kwon*, Dogancan Temel, and Ghassan AlRegib}
\address{Center for Signal and Information Processing,\\ School of Electrical and Computer Engineering,\\ Georgia Institute of Technology, Atlanta, GA, 30332-0250\\ \{mohit.p, gukyeong.kwon, cantemel, alregib\}@gatech.edu}
\begin{document}

\onecolumn 

\begin{description}[labelindent=1cm,leftmargin=3cm,style=multiline]

\item[\textbf{Citation}]{M. Prabhushankar, G. Kwon, D. Temel and G. AIRegib, "Semantically Interpretable and Controllable Filter Sets," 2018 25th IEEE International Conference on Image Processing (ICIP), Athens, 2018, pp. 1053-1057.
} \\

\item[\textbf{DOI}]{\url{https://doi.org/10.1109/ICIP.2018.8451220}} \\

\item[\textbf{Review}]{Date added to IEEE Xplore: 06 September 2018} \\

\item[\textbf{Code/Poster}]{\url{https://ghassanalregib.com/publications/}} \\

\item[\textbf{Bib}] {
@INPROCEEDINGS\{Temel2018\_ICIP,\\ 
author=\{M. Prabhushankar and G. Kwon and D. Temel and G. AIRegib\},\\ 
booktitle=\{2018 25th IEEE International Conference on Image Processing (ICIP)\},\\ 
title=\{Semantically Interpretable and Controllable Filter Sets\},\\ 
year=\{2018\},\\ 
pages=\{1053-1057\},\\ 
doi=\{10.1109/ICIP.2018.8451220\},\\ 
ISSN=\{2381-8549\},\\ 
month=\{Oct\},\}\\
} \\

\item[\textbf{Copyright}]{\textcopyright 2018 IEEE. Personal use of this material is permitted. Permission from IEEE must be obtained for all other uses, in any current or future media, including reprinting/republishing this material for advertising or promotional purposes,
creating new collective works, for resale or redistribution to servers or lists, or reuse of any copyrighted component
of this work in other works. } \\

\item[\textbf{Contact}]{\href{mailto:alregib@gatech.edu}{alregib@gatech.edu}~~~~~~~\url{https://ghassanalregib.com/}\\ \href{mailto:dcantemel@gmail.com}{dcantemel@gmail.com}~~~~~~~\url{http://cantemel.com/}}
\end{description} 

\thispagestyle{empty}
\newpage
\clearpage

\twocolumn

\ninept
\maketitle
\begin{abstract}
In this paper, we generate and control semantically interpretable filters that are directly learned from natural images in an unsupervised fashion. Each semantic filter learns a visually interpretable local structure in conjunction with other filters. The significance of learning these interpretable filter sets is demonstrated on two contrasting applications. The first application is image recognition under progressive decolorization, in which recognition algorithms should be color-insensitive to achieve a robust performance. The second application is image quality assessment where objective methods should be sensitive to color degradations. In the proposed work, the sensitivity and lack thereof are controlled by weighing the semantic filters based on the local structures they represent. To validate the proposed approach, we utilize the CURE-TSR dataset for image recognition and the TID 2013 dataset for image quality assessment. We show that the proposed semantic filter set achieves state-of-the-art performances in both datasets while maintaining its robustness across progressive distortions.    
\end{abstract}
\section{Introduction}
\label{sec:intro}
Visual understanding is a research field that aims to provide semantically meaningful interpretation of the visual cues in data~\cite{Porikli2018}. The objective of visual understanding algorithms is to compute mappings that lead to a representation space in which non-informative and informative representations are distinguishable. Traditional visual understanding algorithms are based on handcrafted approaches in which mappings between input pixel spaces and distinguishable representation spaces are tractable based on data dependent characteristics~\cite{marr1982computational}. Because of the tractable nature of the mapping functions, the representation space spanned by the handcrafted mappings is interpretable. The interpretability and tractability of the representation spaces as well as their mappings have enabled handcrafted visual understanding algorithms to achieve application generalizability. Hence, based on the same underlying representation space, multiple tasks on various applications can be performed using handcrafted feature mappings. For instance, mappings that rely on keypoint detection and feature extraction such as SIFT are used for numerous application including image retrieval~\cite{lowe1999object} and image quality assessment~\cite{temel2016resift}.

Recently, the availability of big data and advancements in computational resources have enabled the development of powerful data-driven algorithms for various vision tasks \cite{ImageNet,krizhevsky2012imagenet}. In data-driven approaches, representation spaces are directly learned from data and labels. A commonly used supervised data-driven method is Convolutional neural network (CNN) which learns a set of discriminative features between classes  ~\cite{krizhevsky2012imagenet}. In this method, the discriminative features are extracted from the input pixel space using a set of learned convolutional filters and mapped to a latent representation space. To interpret the data-driven representation space, the latent representation is visualized in different ways. The authors in~\cite{Zeiler2014} proposed a visualization technique that projects a feature map of interest back to input pixel space. In~\cite{Bau2017}, the authors visualized image patches that maximally activate hidden units and quantified the interpretability of individual filters. These techniques help to interpret the representation space. However, we cannot directly leverage the information obtained from the interpretation because data-driven representations normally contain mixture of abstract features~\cite{olah2017}. Therefore, filters learned in a task-dependent manner are not easily application generalizable and such algorithms require additional training or fine-tuning for different tasks. 


In this paper, we generate controllable semantic filter sets in an unsupervised fashion. The authors in~\cite{yang2016semantic} describe semantic filters as tools to extract subjectively meaningful structures from natural images. We use an autoencoder, an unsupervised neural network, to generate these filters. We demonstrate that, by using such filters, the performance gains of a data-driven approach and the application-generalizability of interpretable models can be combined. The contributions of this paper are threefold:
\vspace{-0.75mm}
\begin{itemize}[leftmargin=*]
\item We analyze various methods to control the training phase of an autoencoder. The filters learned from considered methods are visualized and validated based on their structural interpretability. 
\item We group interpretable filter sets into semantically meaningful visual concepts that are based on color and edge characteristics.
\item We demonstrate the feasibility of semantic filter sets on two contrasting applications including image recognition and image quality assessment. Specifically, we test the robustness of these filters under mild to severe color degradation.
\end{itemize}
\vspace{-0.75mm}
In \cite{temel2016unique} and \cite{prabhushankar2017ms}, we proposed objective image quality estimators based on the representation space spanned by autoencoder filter sets, which were weighed with perceptually inspired formulations. In this paper, we develop further insight into the representation space by delving into the generation of the filter sets to embed semantic meaning within them. In particular, we investigate different regularization techniques for semantically meaningful filter sets and generalize them to applications including recognition under challenging domain shifted conditions. In addition, we analyze the robust performance of image quality estimators under varying challenge levels. Such results provide further insight into understanding the correlation between objective predictors and subjective quality opinions.



\vspace{-0.75mm}
\label{sec:theory}
\begin{figure}[tbh!]
    \centering
    \includegraphics[width=.8\linewidth]{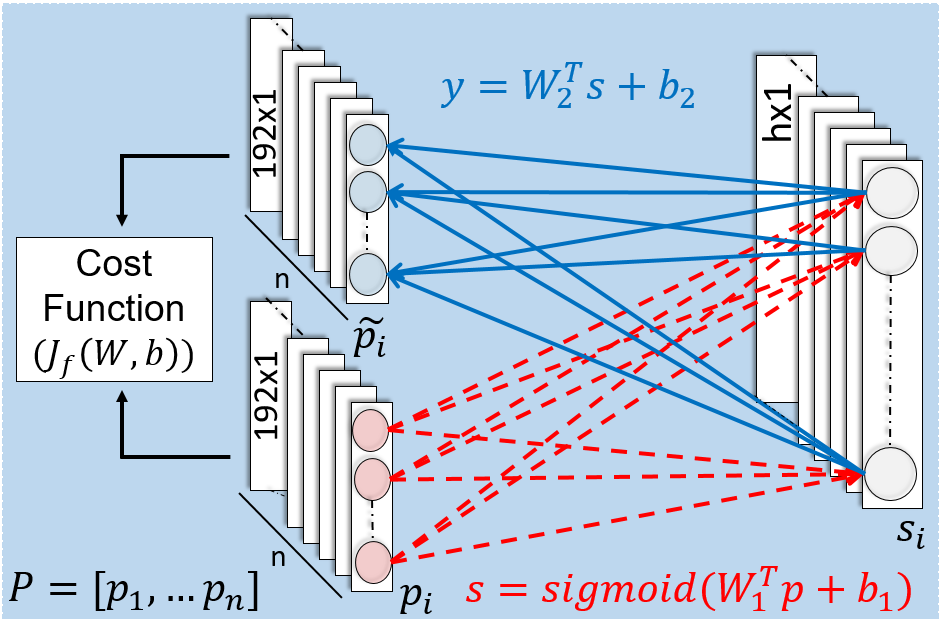}
    \vspace{-0.5mm}
    \caption{Unsupervised training of an autoencoder.}\label{fig:training}
\end{figure}
\vspace{-0.75mm}
\section{Background}
\vspace{-0.75mm}
In this section, we describe a vanilla autoencoder network. An autoencoder is an unsupervised learning network that is trained to copy inputs to outputs~\cite{Goodfellow-et-al-2016}. The network consists of encoder and decoder functions as shown in Fig.~\ref{fig:training}. The encoder maps a matrix $P$, consisting of $n$ input feature vectors each of size $d\times 1$, to a hidden layer with $h$ neurons. The encoder can also be considered as a set of $h$ forward filters each of size $d\times 1$, whose responses are passed through a non-linearity to obtain $s$. Mathematically, $s$ can be represented as,    
\begin{equation}\label{ForwardFilters}
\begin{gathered}
s = \sigma(W_1^T P + b_1),\\
\forall  s\in \Re^{h\times n}, W_1\in \Re^{d\times h}, P\in \Re^{d\times n}, b_1\in \Re^h,
\end{gathered}
\end{equation}
where $s$ is the set of non-linear hidden layer responses to an affine filter set parameterized by weights $W_1$ and bias $b_1$. The sigmoidal non-linearity is used through the rest of this work. The responses $s$ are mapped back to the input data space to obtain $\tilde{P}$ using a decoder. The decoder is a set of backward affine filters parameterized by weights $W_2$ and bias $b_2$. The reconstructed output $\tilde{P}$ is obtained as,
\begin{equation}\label{BackwardFilters}
\tilde{P} = W_2^T s + b_2.
\end{equation}
Note that we use a linear decoder in our experiments. The forward and backward filters are simultaneously trained using backpropagation by constructing and minimizing a cost function $J_f(w,b)$, between $P$ and $\tilde{P}$ given by,
\begin{equation}\label{RegularizedBackprop}
J_f(W,b) = \lVert W_2^T(\sigma(W_1^T P + b_1)) + b_2 - P \rVert_2^2 + f(W,b),
\end{equation}
where the first term is the Mean Square Error (MSE) between the input and reconstructed outputs, and $f(W,b)$ is a regularization term. Regularization is a modification of an optimization objective function to reduce the generalization error without impacting the training error~\cite{Goodfellow-et-al-2016}. Practically, adding regularization to an autoencoder means that the network is restricted to copying only an approximate input as its output. This forces the autoencoder to prioritize the necessary aspects of the input thereby learning a dictionary of useful properties that characterize the input data. The need for regularization is well recognized within the machine learning community~\cite{bauer2007regularization} and a number of regularization techniques have been proposed. The commonly used techniques for autoencoders include weight penalties, derivative penalties, and training tricks including early stopping~\cite{yao2007early}. In this work, we concentrate on analyzing weight penalties. This is in keeping with the overall theme of generating semantic filters that are parameterized by weights. While these techniques have already been proposed, our focus is to to control the semantic meaning and visual concepts learned by the weights.
\vspace{-0.5mm}
\begin{figure}[tbh!]
    \centering
    \includegraphics[width=0.8\linewidth]{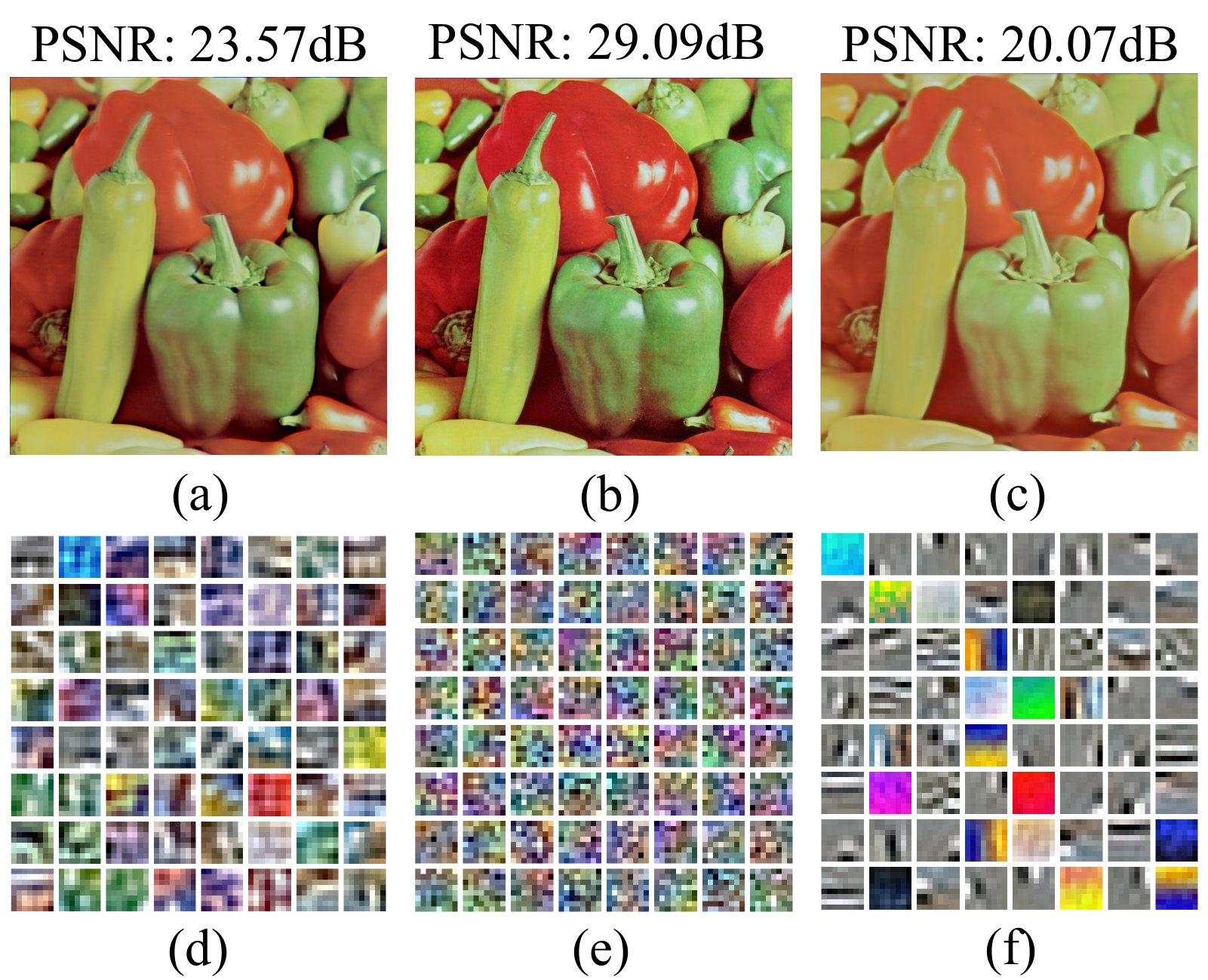}
    \vspace{-0.5mm}
    \caption{(a), (b), (c): Reconstructed images from $l1$, $l2$, and elastic net regularization, respectively, (e), (f), (g): Weights of autoencoder from $l1$, $l2$, and elastic net regularization, respectively.}\label{fig:reconstruction}
\end{figure}
\vspace{-0.5cm}
\vspace{-0.75mm}
\section{Regularization analysis}
\label{sec:analysis}
We train an autoencoder on ImageNet database~\cite{ImageNet} to obtain encoder filters. 100 patches of dimension $8 \times 8 \times 3$ are sampled from each of $1,000$ images and preprocessed using Zero-Phase Component Analysis (ZCA) whitening~\cite{ZCA}. These whitened patches are fed into multiple autoencoders, each with different weight penalties, $f(W,b)$ from eq.~\ref{RegularizedBackprop}. The different penalties are $\ell_1$ or LASSO~\cite{tikhonov2013numerical}, $\ell_2$ or ridge regression~\cite{tibshirani1996regression}, and elastic net~\cite{zou2005regularization} regularization which is a weighted combination of $\ell_1$ and $\ell_2$ penalties. While all these penalties have been explored in regularization theory, we provide a qualitative analysis based on the visual concepts that each of the regularization techniques learns. 
\vspace{-0.75mm}
\subsection{\texorpdfstring{$\ell_1$}{} penalty}
\label{sec:L1}
\vspace{-0.75mm}
The $\ell_1$ constraint promotes sparsity within the hidden layer responses $s$. The training cost function for $\ell_1$ regularization is given by,
\begin{equation}\label{eq:L1}
J_{\ell_1}(W,b) = \lVert W_2^T(\sigma(W_1^T P + b_1)) + b_2 - P \rVert_2^2 + \beta\lVert W \rVert_1.
\end{equation}
where $\beta$ is a positive regularization parameter. The trained filters are visualized in Fig.~\ref{fig:reconstruction}(d). An example peppers image is passed through the autoencoder and its reconstructed image is visualized in Fig.~\ref{fig:reconstruction}(a). The fidelity of the reconstructed image is $23.57$ dB. 

Theoretically, $\ell_1$ regularization can help with interpretability since, for every train and test case, only a few filters are activated. However, in practice the $\ell_1$ penalty suffers when there are correlated weights or filters. If there is a group of filters among which the pairwise correlations are very high, then the $\ell_1$ penalty selects any one of those correlated filters~\cite{zou2005regularization}. This filter selection issue is not suitable for applications where the representation space is used as features for further analysis. Filters learned from images have large correlations between them because of the inherent spatial correlation within training data. Also, as shown in Fig.~\ref{fig:reconstruction}(d), the interpretability of the mapping functions is not obvious.
\subsection{\texorpdfstring{$\ell_2$}{} penalty}
\label{sec:L2}
The $\ell_2$ penalty in the cost function is also called ridge regression, weight decay or Tikhnov regularization~\cite{tikhonov2013numerical}. This is a well studied regularization technique that promotes shrinkage between filters. The cost function with $\ell_2$ penalty on weights is given by, 
\begin{equation}
J_{\ell_2}(W,b) = \lVert W_2^T(\sigma(W_1^T P + b_1)) + b_2 - P \rVert_2^2 + \lambda\lVert W \rVert_2^2.
\end{equation} 
By penalizing the filters in this manner, the input data produces dense responses. However, a mixture of large and small responses causes instability while learning~\cite{zou2005regularization}. Practically, this means that the representation space changes rapidly even with slight domain shift between train and test data. For application-generalizability, the $\ell_2$ regularization may not be suitable. The filter sets learned from this technique along with the reconstructed peppers image are provided in Fig.~\ref{fig:reconstruction}(b) and ~\ref{fig:reconstruction}(e) respectively. Again, the filters are able to reconstruct the image with a PSNR score of $29.09$ dB. However, there is no semantic interpretability that can be associated with either the filters or the activations.
\subsection{Elastic Net penalty}
\label{sec:ElasticNet}
The elastic net was proposed in~\cite{zou2005regularization} to correct for the filter selection challenge faced by $\ell_1$ penalty. Instead of one filter being activated from a group of correlated filters, all the correlated filters are activated at the same time. This is achieved by adding a weighted $\ell_2$ norm penalty to obtain within-group dense responses. A thorough explanation and analysis is provided in~\cite{zou2005regularization}. The minimization cost function for this regularization technique is given by,
\begin{equation}\
J(W,b) = \lVert\tilde{P} - P \rVert_2^2 + \beta\lVert W \rVert_1 + \lambda\lVert W \rVert_2^2.
\end{equation}
The values of $\beta$ and $\lambda$ are set to $5$ and $3e^{-3}$ respectively as suggested by the authors in \cite{ng2012ufldl}. The weights learned using the elastic net regularization along with the reconstructed peppers image are visualized in Fig.~\ref{fig:reconstruction}(c) and ~\ref{fig:reconstruction}(f) respectively. The lower fidelity of the reconstructed image is expected because of the higher regularization penalties. However, it is obvious that each filter in Fig.~\ref{fig:reconstruction}(c) has a \textit{visual concept} associated with it. These concepts include different colors, color gradients, and edges with multiple orientations. These weights are structurally meaningful and semantic in nature. While other works in sparse coding~\cite{ZCA} and CNNs have shown that edges dominate the first layer with just $\ell_2$ penalty, the complete demarcation between color and edge filters using elastic net penalty is very interesting and deems further investigation. 
\vspace{-1.5mm}
\section{Semantic Visual Concepts}
\label{sec:generation}
\vspace{-1.5mm}
To leverage these semantic filters for different tasks, we further group them based on individual visual concepts. The grouping of semantic filter sets is performed based on the kurtosis measure of each filter. Kurtosis, $\kappa$, is  defined as,
\begin{equation}
    \kappa[W_1] = \frac{E[W_1-\mu]^4}{\sigma^4},
\end{equation}
where $W_1$ are vectorized, zero-centered and normalized encoder filter values, $\mu$ and $\sigma$ are respectively mean and the standard deviation of $W_1$. For every $\kappa[W_1]$ greater than a threshold of $5$, the filter is classified as an edge filter and if $\kappa[W_1] < 2$, the filter is classified as a color filter. The thresholds are empirically determined to ensure a complete demarcation between filters that represent color and edges. Edge filters have a higher kurtosis value because they have values in a localized area. Therefore, most of the filter values are located away from the mean of distribution leading to higher kurtosis. While the authors in~\cite{majumdar2009classification} and~\cite{ng2011sparse} use elastic net regularization for learning, they do not go further to group filters by their representative visual concepts. We note that the entire process from obtaining interpretable filters to grouping semantic filters is conducted in an unsupervised fashion. 
\section{Experiments and Results}
\label{sec:application}
\begin{figure}[tbh!]
    \centering
    \includegraphics[width=.95\linewidth]{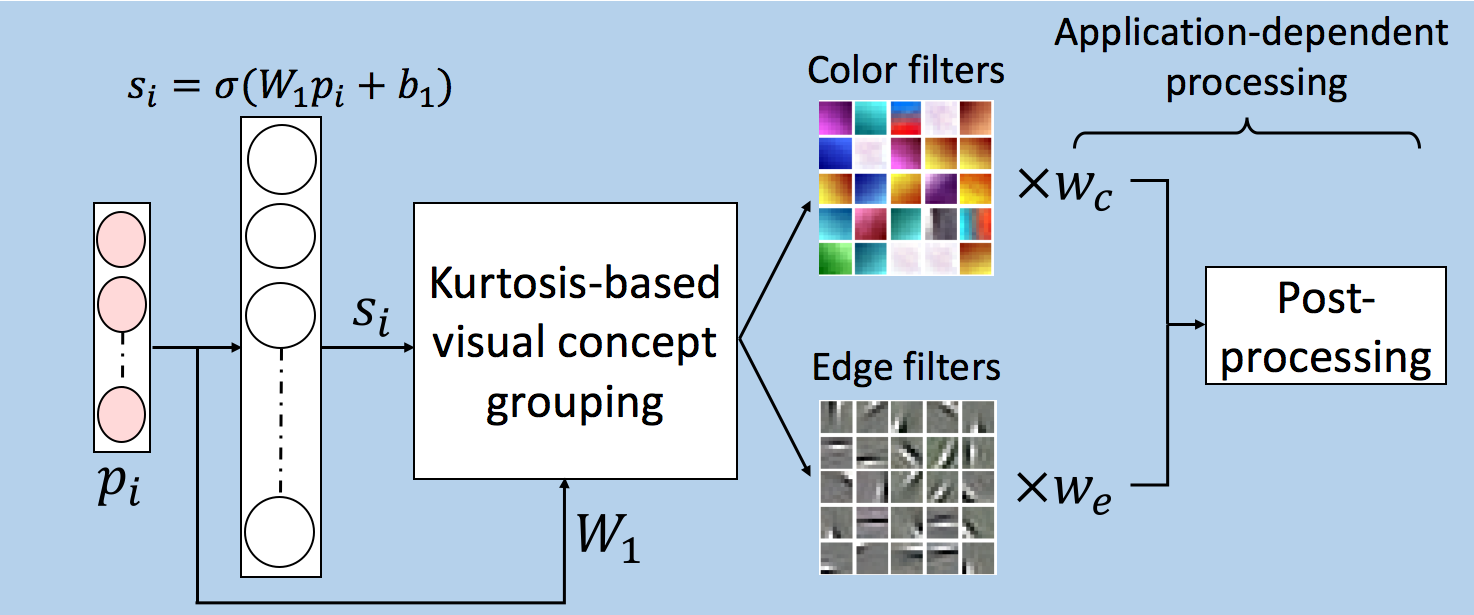}
    \vspace{-0.2cm}
    \caption{General framework of the semantic autoencoder.}\label{fig:testing}
\end{figure}

In this section, we demonstrate the advantages of learning semantic filters and categorizing them based on visual concepts with two contrasting applications:
\begin{itemize}[leftmargin=*]
\item Recognition under progressive decolorization challenge where the objective is to recognize decolorized images when trained on color images of the same class.
\item Image Quality Assessment (IQA) under color distortion where the goal is to objectively estimate the subjective scores of images affected by color distortion.
\end{itemize}
In these two applications, we show proposed semantic filter-based methods outperform state-of-the-art algorithms while maintaining its robust performance under different levels of color distortion. For validation, we use the Challenging Unreal and Real Environments for Traffic Sign Recognition (CURE-TSR) dataset~\cite{Temel2017} and the TID 2013~\cite{ponomarenko2015image} dataset, for recognition and IQA respectively. They contain images with decolorization and color distortion challenges across $5$ progressive levels. After the semantic filters are grouped based on the visual concepts, we prune the filters in an application dependent manner. Pruning is carried out by applying application-dependent weights $(w_c, w_e)$ on color and edge filter sets. We refer to this model as a semantic autoencoder (Sem-AE) and a general framework is illustrated in~\ref{fig:testing}. The $p_i$ refers to the whitened patches extracted from images. The filter set $W_1$ is grouped into visual concepts and their responses are processed in an application-dependent manner.
\vspace{-0.5mm}
\vspace{-0.5cm}

\subsection{Recognition}
\vspace{-0.5mm}
\begin{figure}[tbh!]
    \centering
    \includegraphics[width=0.8\linewidth]{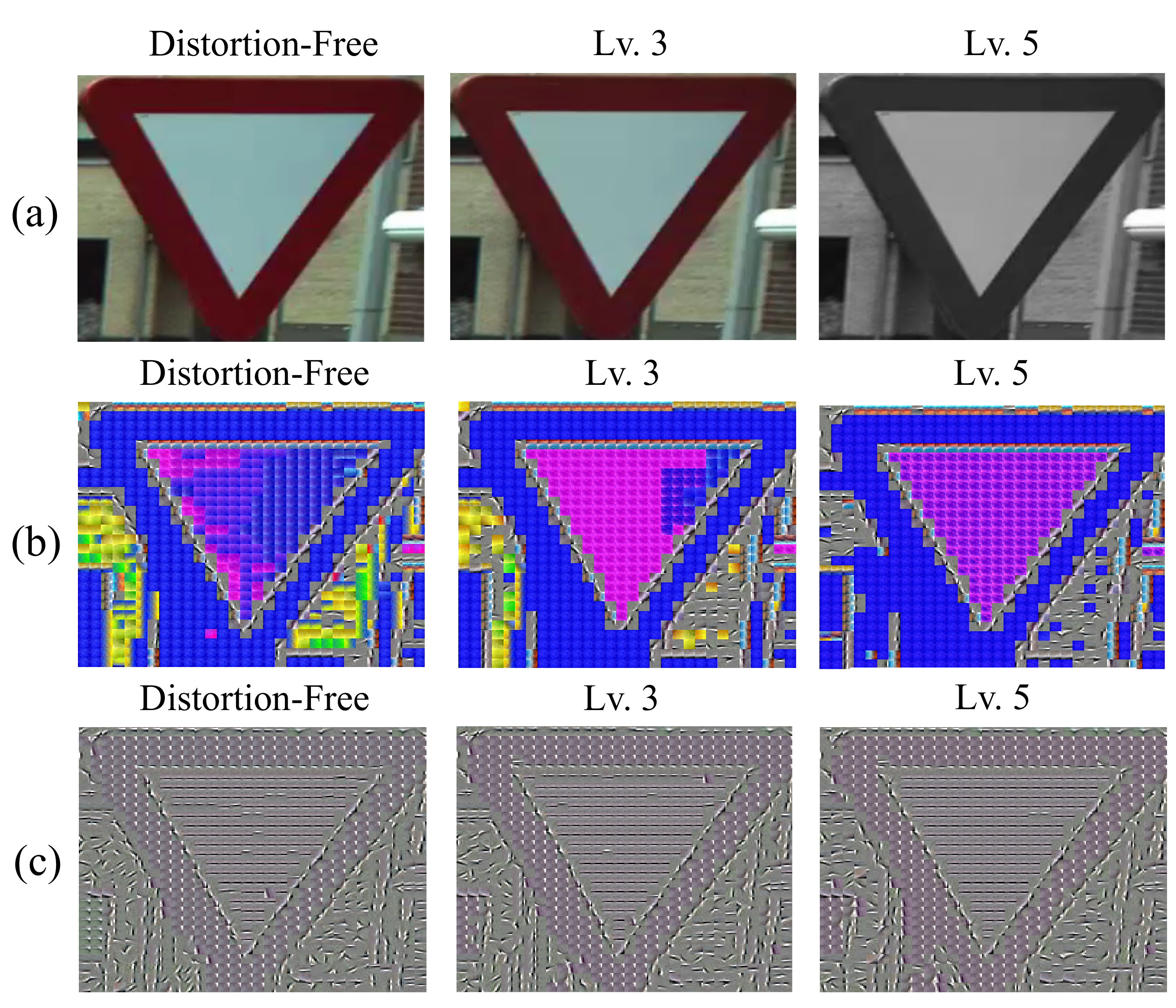}
    \vspace{-0.5mm}
    \caption{Visualization of filters that are maximally activated by each patch in the \texttt{yield} traffic sign image. (1st row: raw images, 2nd row: the maximally activated filters from both edge and color filters, 3rd row: the maximally activated filters from edge filters.)}\label{fig:tsr_visualization}
    \vspace{-0.35cm}
\end{figure}
\begin{figure}[tbh!]
    \centering
    \includegraphics[width=0.8\linewidth]{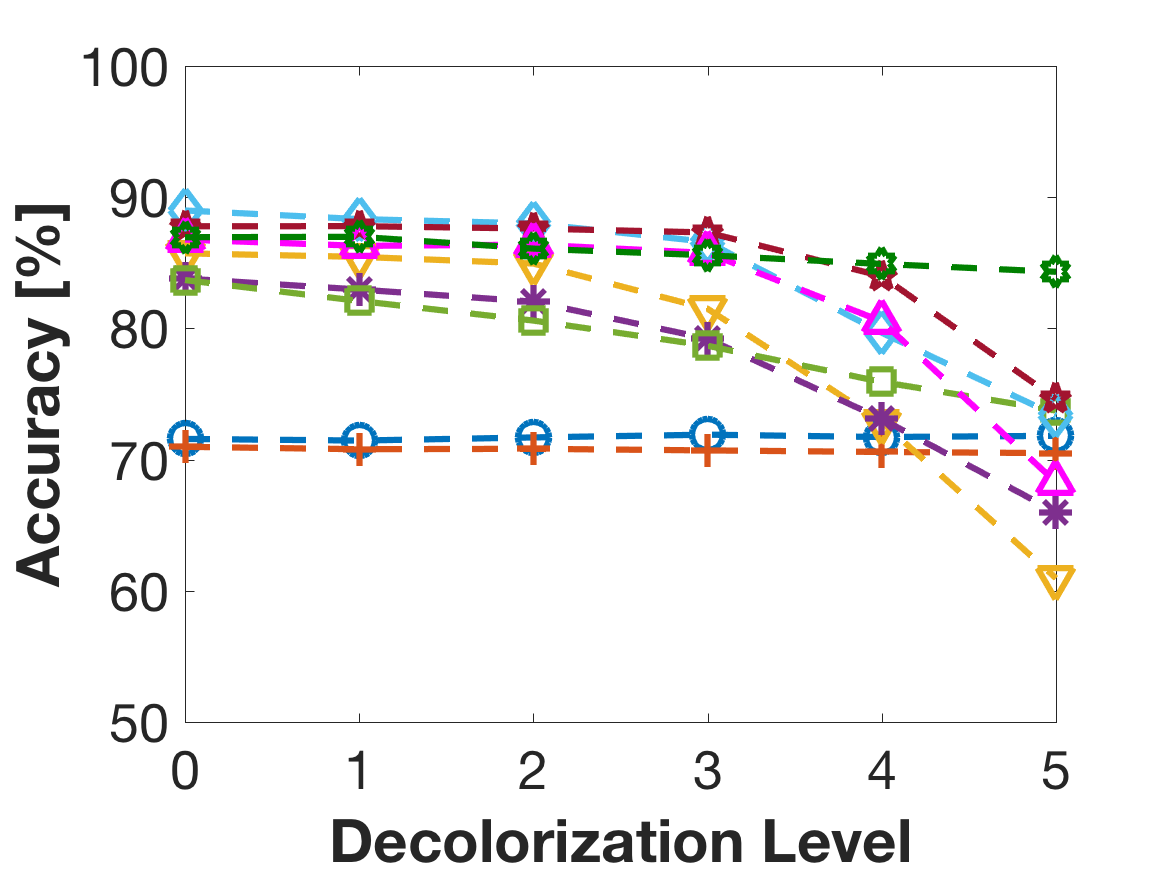}
    \includegraphics[width=\linewidth]{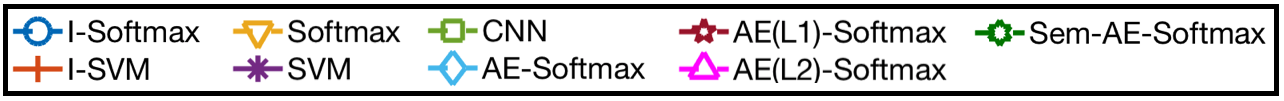}
    \vspace{-0.5cm}
    \caption{Accuracy of traffic sign recognition for different levels of decoloriazation.}\label{fig:tsr_acc}
\end{figure}

The CURE-TSR dataset consists of $7,292$ challenge-free traffic sign images over $14$ classes for training. Testing is performed on $20,004$ images that includes $3,334$ images in each of distortion-free and five different levels of decolorization category (total $3,334 \times 6 = 20,004$ images). An example of distortion-free, decolorization levels $3$ and $5$ are shown in the Fig.~\ref{fig:tsr_visualization}(a). In the Fig.~\ref{fig:tsr_visualization}(b), we visualize filters which show maximum activations for each patch in distortion-free, level $3$, and level $5$ decolorization images among all $W_1$ filters including color and edge concepts. In the Fig.~\ref{fig:tsr_visualization}(c), only the edge concept filters are used for visualization. It is apparent that the representation obtained only from edge filters are not adversely affected by decolorization while the representation obtained from all the filters change significantly. Hence, we assign $(w_c, w_e)$ to be $(0, 1)$ and only the edge filters are used to obtain features. A softmax classifier is trained on these edge features as part of the application-dependent post processing block in Fig.~\ref{fig:testing}. Performance accuracies of the Sem-AE-based recognition algorithm across decolorization levels are plotted in Fig.~\ref{fig:tsr_acc}. We compare proposed algorithm with other baseline algorithms detailed in~\cite{Temel2017}. In addition, we show performance of $l_1$ (AE(L1)), $l_2$ (AE(L2)), and elastic net (AE) regularized filters on which softmax classifiers are trained. The performance of two intensity-based methods (I-Softmax, I-SVM) is not affected by decolorization since they do not use color information. However, by discarding color information, these methods achieve lower performance among the first $4$ levels than all the other methods. It can be seen that the Sem-AE with $(w_c, w_e) = (0, 1)$ shows high and steady accuracy across decolorization levels while the accuracy of all other RGB based methods degrades as decolorzation becomes severe. The steady performance indicates that the representation space spanned by Sem-AE filters is robust to color distortions, because the edge-based filter responses are invariant to color degradations. 
\vspace{-0.3cm}
\vspace{-0.5cm}
\subsection{Image Quality Assessment}
For the task of Image Quality Assessment (IQA), we use TID 2013 database to evaluate the performance of Sem-AE. In particular, we analyze the categories of color saturation, color quantization with dither and chromatic aberrations. We choose these categories to specifically illustrate the worth of learning color and edge concepts for IQA. Each category has $5$ levels of progressive color based distortions. Previous studies ~\cite{Ponomarenko2011,Wang2003,Zhang2011,Zhang2012,Temel2015} in this field, including our own \cite{temel2016unique,prabhushankar2017ms} concentrated on and provided results on all the distortion levels together. In this section, we report results individually for each level. Ideally, a good quality estimator is correlated with subjective scores across varying distortions. 

The Sem-AE scores are obtained for both reference and distorted images with $(w_c, w_e) = (0.5, 2)$. The higher weighting is given to edges because the human visual system is more sensitive to sharpness of images and distortion in the edge components of images causes higher visual discomfort~\cite{Hassen2010}. Spearman correlation is calculated between the weighted responses of distorted and reference images to obtain the objective scores. In Table~\ref{table:pcc_scc}, we compare (Sem-AE) with other commonly compared metrics. Pearson (PCCs) and Spearman Correlation Coefficients (SCCs) which measure linearity and monotonic behavior between subjective and objective image quality scores are used to validate the objective metrics. We observe that Sem-AE follows the subjective scores closer than the other metrics. Not only does it exhibit the highest correlation in $4$ of the $5$ categories in both PCC and SCC, it maintains its steady performance across levels. This is in contrast to other methods which show low correlations in levels $1-3$. Note that both FSIMc and PerSIM take color characteristics into consideration. The tested quality estimators exhibit uneven correlations across distortion levels, which should be addressed to obtain robust methods.

\begin{table}[thb!]
\centering
\caption{Pearson correlation coefficients and Spearman correlation coefficients for different decolorization level}
\label{table:pcc_scc}
\begin{tabular}{|c| c c c c c|}
\hline
\hline
\multirow{2}{*}{Metric} & \multicolumn{5}{c|}{Pearson Correlation Coefficient} \\ \cline{2-6} 
                        & Lv. 1      & Lv. 2      & Lv. 3     & Lv. 4     & Lv. 5     \\ \hline
PSNR-HMA                & 0.643      & 0.626      & 0.280     & 0.046     & 0.486     \\ \hline
MS-SSIM                 & 0.248      & 0.143      & 0.302     & 0.525     & 0.744     \\ \hline
SR-SIM                  & 0.370      & 0.260      & 0.301     & 0.497     & 0.732     \\ \hline
FSIMc                   & 0.391      & 0.253      & 0.303     & 0.553     & 0.778     \\ \hline
PerSIM                  & 0.126      & 0.085      & 0.304     & 0.554     & \textbf{0.804}     \\ \hline
AE                 & 0.716      & 0.725      & 0.765     & 0.775     & 0.577     \\ \hline
AE(L1)                  & 0.557      & 0.406      & 0.542     & 0.682     & 0.619     \\ \hline
AE(L2)                  & 0.079      & 0.004      & 0.275     & 0.454     & 0.568     \\ \hline
Sem-AE        & \textbf{0.772}      & \textbf{0.795}      & \textbf{0.801}     & \textbf{0.816}     & 0.730     \\
\hline\hline
\multirow{2}{*}{Metric} & \multicolumn{5}{c|}{Spearman Correlation Coeffcieint}                        \\ \cline{2-6} 
                        & Lv. 1          & Lv. 2          & Lv. 3          & Lv. 4          & Lv. 5          \\ \hline
PSNR-HMA                & 0.505          & 0.475          & 0.140          & 0.229          & 0.732          \\ \hline
MS-SSIM                 & 0.471          & 0.345          & 0.111          & 0.224          & 0.691          \\ \hline
SR-SIM                  & 0.505          & 0.401          & 0.098          & 0.234          & 0.732          \\ \hline
FSIMc                   & 0.432          & 0.347          & 0.013          & 0.395          & 0.793          \\ \hline
PerSIM                  & 0.306          & 0.160          & 0.143          & 0.479          & \textbf{0.825} \\ \hline
AE                 & 0.648          & 0.764          & 0.795          & 0.786          & 0.389          \\ \hline

AE(L1)                  & 0.451          & 0.378          & 0.541          & 0.660          & 0.480          \\ \hline
AE(L2)                  & 0.084          & 0.120          & 0.188          & 0.381          & 0.543          \\ \hline
Sem-AE        & \textbf{0.725} & \textbf{0.815} & \textbf{0.802} & \textbf{0.797} & 0.615          \\ \hline\hline
\end{tabular}
\end{table}
\vspace{-0.5cm}


\section{Conclusion}
In this paper, we analyzed existing regularization techniques to obtain semantic filter sets using an unsupervised learning technique. The semantic filters which represent characteristic visual concepts were learned jointly from natural images. While the learned semantic filters succeeded in their primary task of reconstructing an input image, their worth was illustrated in two applications that employed their color and structural groupings. This work provides a promising step towards defining perceptual visual concepts that can be used to learn, interpret, and leverage deep learning models.

\end{document}